%% file: main.tex
\documentclass[runningheads]{llncs}

\usepackage[T1]{fontenc}
\usepackage{amsmath}
\usepackage{amsfonts}       
\usepackage{caption}
\usepackage{hyperref}
\usepackage{graphicx,verbatim}

\usepackage{color}

\usepackage{booktabs}       
\usepackage{multirow}       
\usepackage{pifont}         

\newcommand{\cmark}{\ding{51}}%
\newcommand{\xmark}{\ding{55}}%

\begin{document}
\title{Random Direct Preference Optimization for Radiography Report Generation}
\titlerunning{RDPO for Radiography Report Generation}
%
\author{Valentin Samokhin\inst{1,3} \and
        Boris Shirokikh\inst{1,2}$^{\text{,}*}$ \and
        Mikhail Goncharov\inst{1,2} \and
        Dmitriy Umerenkov\inst{1,2} \and
        Maksim Bobrin\inst{1,2} \and
        Ivan Oseledets\inst{1,2} \and
        Dmitry Dylov\inst{1,2} \and
        Mikhail Belyaev\inst{4}
}
\authorrunning{V. Samokhin et al.}
\institute{ Artificial Intelligence Research Institute (AIRI), Moscow, Russia \and
            Skolkovo Institute of Science and Technology, Moscow, Russia \and
            Institute for Information Transmission Problems, Moscow, Russia \and
            Independent researcher \\
            $^*$ Correspondence \email{boris.shirokikh@skoltech.ru}
}



\maketitle              

\input{sections/0_abstract}

\input{sections/1_intro}

\input{sections/4_related_work}
\input{sections/2_method}

\input{sections/3_experiments}

\input{sections/5_conclusion}

%
%
%

\bibliographystyle{splncs04}
\bibliography{main.bib}
%




\end{document}

%% file: sections/0_abstract.tex
\begin{abstract}
    Radiography Report Generation (RRG) has gained significant attention in medical image analysis as a promising tool for alleviating the growing workload of radiologists. However, despite numerous advancements, existing methods have yet to achieve the quality required for deployment in real-world clinical settings. Meanwhile, large Visual Language Models (VLMs) have demonstrated remarkable progress in the general domain by adopting training strategies originally designed for Large Language Models (LLMs), such as alignment techniques. In this paper, we introduce a model-agnostic framework to enhance RRG accuracy using Direct Preference Optimization (DPO). Our approach leverages random contrastive sampling to construct training pairs, eliminating the need for reward models or human preference annotations. Experiments on supplementing three state-of-the-art models with our Random DPO show that our method improves clinical performance metrics by up to 5\%, without requiring any additional training data.

    \keywords{Radiography \and Report Generation  \and Visual Language Models \and Alignment \and Direct Preference Optimization.}

\end{abstract}

%% file: sections/1_intro.tex
\section{Introduction}
\label{sec:intro}

Chest radiography (CXR) has become the global standard for the non-invasive assessment of major thoracic organs. It is a fast and a widely used screening tool that provides valuable diagnostic insights while minimizing radiation exposure. CXR plays a crucial role in detecting both acute and chronic conditions affecting the lungs, great vessels, mediastinum, and osseous structures \cite{ko2024demands}.

The growing global demand for CXR examinations has placed an increasing burden on radiologists responsible for interpreting these studies. And the number of radiographic studies is rising faster than the number of radiologists available to review them \cite{ko2024demands}. This increasing workload heightens the risk of diagnostic errors due to fatigue and time constraints. In the context of thoracic pathologies, one of the leading causes of mortality, such errors can have severe consequences.

Automated methods for radiology report generation (RRG), particularly those based on large Visual Language Models (VLMs) \cite{bannur2024maira,chexpert_plus,cvt2distilgpt2}, offer a promising solution to alleviating the increasing burden on radiologists. These systems can serve as a second opinion, assisting in the detection of major pathologies and reducing the likelihood of missed findings. However, despite recent advancements, even state-of-the-art solutions fall short of the quality required for deployment in real-world clinical settings \cite{xu2024overview,zhang2024rexrank}.



Recent studies \cite{bannur2024maira,nicolson2024health,chen2024chexagent} have shown that incorporating auxiliary tasks, such as grounding or visual question answering, into the training pipeline enhances model performance on RRG benchmarks. This suggests that the conventional next-token prediction objective, commonly used for radiograph captioning, is insufficient for learning clinically accurate reports. One of the key challenges here is the inherent variability in clinical reports -- they are written by different physicians, each with their own preferred style and terminology.

The latter problem can be approached using the general domain methods, where VLMs and Large Language Models (LLMs) ensure their outputs align with highly-varying human preferences. Direct Preference Optimization (DPO) \cite{dpo} is one widely adopted method for this: it refines model responses by comparing two alternatives, a chosen and a rejected response, and optimizing the model to favor the preferred one. This preference-based learning paradigm has been shown to improve output quality in various language and vision-language tasks, suggesting its potential for enhancing RRG performance as well.

Given the high variability in clinical guidelines and reporting styles, we assume that even randomly sampling a rejected response from the same dataset could positively impact model performance after DPO. Herein, we refer to this approach as Random DPO (RDPO). We introduce it as a universal and model-agnostic training framework that requires only a dataset of ground-truth image-text pairs and a pretrained RRG model checkpoint, eliminating the need for reward models, human preference annotations, or any extra training data.

We empirically show how RDPO improves three publicly available models on the largest relevant datasets, such as MIMIC-CXR \cite{mimic} and CheXpert Plus \cite{chexpert_plus}. The gain it provides is consistent and is up to 5\% in terms of clinically relevant metrics. We summarize our contributions as follows:

\let\labelitemi\labelitemii
\begin{itemize}
    \item We propose Random DPO, a framework for enhancing RRG models, without requiring additional labels, rewards, or preference annotations.
    \item We set a new state-of-the-art result by taking the strongest publicly available checkpoint MAIRA-2 \cite{bannur2024maira} and applying our method on top of it.
\end{itemize}











%% file: sections/4_related_work.tex
\subsubsection{Related work}
\label{sec:related_work}

Over the years, classification and segmentation tasks have been effectively addressed by deep learning techniques, showing robust performance in identifying various pathologies \cite{meedeniya2022chest}. However, the automatic generation of textual radiology reports remains largely unsolved, despite recent progress in assessing report quality \cite{yu2023evaluating}. Emerging vision-language models, such as MAIRA-2 \cite{bannur2024maira} and CheXagent \cite{chen2024chexagent}, demonstrate significant promise in bridging the gap between image analysis and report generation, yet they are still not sufficiently optimized to meet the rigorous standards required for widespread clinical adoption.

Early deep learning (DL)-based approaches used convolutional neural networks (CNNs) to generate radiology reports \cite{gale2018producing}. Next, recurrent neural network (RNN) methods, including a hierarchical CNN-RNN model described in \cite{liu2019clinically}, advanced chest X-ray report generation. Attention-enhanced RNN architectures further refined outputs by focusing on clinically relevant features \cite{xue2018multimodal}. 
Finally, large-scale vision-language models (VLMs), which integrate specialized image encoders and large language models (LLMs) via a trained encoder, show promising results \cite{bannur2024maira}, \cite{chen2024chexagent}.


%% file: sections/2_method.tex
\section{Method}
\label{sec:method}

\subsection{Background}

From an objective standpoint, RRG models are typically optimized using standard cross-entropy loss for next-token prediction \cite{iu_xray}. This naturally encourages the model to follow an auto-regressive decoding process, generating the ground-truth caption step by step \cite{Selivanov2023}. 
However, in practice, this choice of objective function introduces unintended biases. 
The model may learn common text patterns that frequently co-occur and erroneously associate them with visual patterns that are not necessarily related to the text. These biases arise due to both the limitations of the training data (such as its quantity and diversity of sources), as well as domain shifts in clinical data distributions.

In general domain, Large Language Model training is not limited by supervised fine-tuning (SFT) on a narrow dataset. Instead, it is followed by the alignment step, which \textit{aligns} model's responses with the human values.
Among many alternatives, Direct Preference Optimization \cite{dpo} is considered to be the most simple to reproduce. It allows one to get rid of an extra reward model and directly use preference dataset. In addition to input prompts, it consists of both chosen, or preferred, responses and rejected, or negative, responses. It does not imply that rejected response is an incorrect answer to the given prompt, but it means that according to some subjective criteria this specific option is less preferable. 


In our task, fully identical cases are extremely rare — at most, they differ in minor details or wording. This assumption, supported by empirical evidence, motivates experiments where random samples are treated as rejected ones. Surprisingly, this approach leads to quality improvements across multiple metrics, regardless of the pretrained checkpoints used as a starting point or the test datasets evaluated.

\subsection{Formal description}

We denote starting model checkpoint as $\pi_{ref}$, and optimized model, initialized from this checkpoint, as $\pi_{\theta}$.
At each training step, we sample a batch of image-text pairs of size $N$: $X = \{(x_i, y_i)\}_{i=1}^N$ from the training set. To construct pairs for preference optimization, for each pair $(x_i, y_i)$ we sample a random $y_{j_i}$ from $\{y_k\}_{k=1}^N$, different from $y_i$. We can rename $y_i$ as $y^w_i$ - chosen response, and $y_{j_i}$ as $y^l_i$ - rejected response, and get DPO loss:

\begin{equation}
\mathcal{L}(\theta)= -\mathbb{E}_{_{X \sim D}}
\left[\sum_{i=1}^N
\log \sigma
\left(
\beta \log \frac{\pi_{\theta}(y^w_i \mid x_i)}{\pi_{ref}(y^w_i \mid x_i)}
- \beta \log \frac{\pi_{\theta}(y^l_i \mid x_i)}{\pi_{ref}(y^l_i \mid x_i)}
\right)
\right].
\end{equation}

In our implementation, we do not need to prepare response pairs in advance. Instead, we sample pairs at the batch collation step without need of allocating extra space. Our approach does not depend on choice of pretrained model $\pi$, source of training data $D$, or alternative Preference Optimization algorithm. Thus, the framework can be modified according to one's needs. Importantly, the framework does not require any extra data preparation or annotation and simply leverages datasets already collected for SFT, image captioning, or fine-tuning.







%% file: sections/3_experiments.tex
\section{Experiments}

\subsection{Datasets}
\label{ssec:datasets}

In our experiments, we use four publicly available datasets that provide CXR images paired with radiological reports: MIMIC-CXR \cite{mimic}, CheXpert Plus \cite{chexpert_plus}, IU X-ray \cite{iu_xray} and Interpret-CXR. These datasets were collected from diverse clinical settings, enabling comprehensive evaluation of the RRG methods.

\textbf{MIMIC-CXR} contains 377,110 images corresponding to 227,835 radiographic studies. These studies are sourced from the emergency department of the Beth Israel Deaconess Medical Center in Boston, MA, and each study is accompanied by a corresponding radiological report.

\textbf{CheXpert Plus} comprises 223,228 unique pairs of radiology reports and chest X-rays, derived from 187,711 studies for 64,725 patient, acquired in both inpatient and outpatient centers. The curation and release of CheXpert Plus were supported in part by the Medical Imaging and Data Resource Center (MIDRC).

\textbf{IU X-Ray} is a publicly available dataset containing 7,470 pairs of radiology reports and chest X-rays, collected by Indiana University.

\textbf{Interpret-CXR} is a composite dataset that integrates MIMIC-CXR, CheXpert Plus, PadChest~\cite{padchest}, and BIMCV COVID-19~\cite{bimcv}. PadChest contributes 104,393 X-ray studies, interpreted and reported by radiologists at the Hospital San Juan in Spain between 2009 and 2017. BIMCV COVID-19 adds 46,727 chest X-ray image-text pairs from the Data Bank of Medical Images of the Valencian Community. Since Interpret-CXR does not have a publicly available test set, we use it exclusively for alignment.

MIMIC-CXR and CheXpert Plus are used in both RDPO alignment and evaluation, while IU X-Ray is left for evaluation purposes due to its small size. For all datasets, we use the officially published train-val-test splits, ensuring consistency and reproducibility of our experiments.

\subsection{Metrics}

\input{tables/results_1_mimic}

In this work, we use five metrics: BLEU \cite{papineni2002bleu}, Bert-score \cite{zhang2019bertscore}, Semb-score \cite{endo2021retrieval}, Radgraph-combined \cite{jain2021radgraph} and RadCliQ-v1 \cite{yu2023evaluating}. BLEU remains a widely accepted n-gram-based metric that measures lexical overlap to provide a rough assessment of linguistic similarity. Semb-score evaluates semantic alignment at the concept level by comparing CheXbert's embeddings, emphasizing the preservation of meaning within the generated text. Radgraph-combined integrates graph-based representations of radiological entities and their relationships, offering a holistic view of both factual correctness and contextual coherence. RadCliQ-v1 is a composite metric created to be correlated with experts' preferences. For the consistency with other metrics (\textit{higher is better}), we report inverted RadCliQ-v1.

\subsection{Experimental design}

The starting point for our experiments are model checkpoints, trained on image-captioning task in the SFT manner. To make experiments fair, for primary tests we chose models, trained only on public RRG datasets on RRG task only. This strategy allows us to use the same training data and eliminate negative effect of potential catastrophic forgetting.

\subsubsection{Stanford AIMI's models from Chexpert-Plus release}\cite{chexpert_plus} - a vision encoder-decoder model. Encoder is Swinv2 \cite{swinv2}, decoder is BertDecoder \cite{bert-decoder} with 2 layers. We do experiments with three checkpoints, trained on combinations MIMIC-CXR and Chexpert-Plus.

\subsubsection{Cvt2Distilgpt2 (CSIRO)}  \cite{cvt2distilgpt2} -  a vision encoder-decoder model. There are two checkpoints publicly available - trained on MIMIC and IU-Xray, respectively. We omit the IU-Xray checkpoint due to the size of training set insufficient for DPO. Encoder was initialized from Convolutional Transformer \cite{cvt}, while decoder was initialized from DistilGPT-2 \cite{distilgpt2}

\subsubsection{MAIRA-2} As a proof-of-concept, we tried to replicate our results on the strongest publicly available RRG model, MAIRA-2\cite{bannur2024maira}, despite that it was partially trained on private data. It is a LLaVa-like model \cite{llava}, trained on multiple datasets and tasks, demonstrates superior performance across multiple benchmarks. Vision encoder is DINOv2 \cite{dinov2} pretrained on mix of public and private radiographs. LLM decoder is Vicuna-7b-v1.5. 

\subsubsection{Training hyperparameters} We conducted all fine-tuning experiments using LoRA \cite{lora} with $\alpha=16$ and $r=8$ and over $q$ and $v$ weights of attention layers. Learning rate was set to $1e-6$ with cosine scheduler, batch size to 64, and number of training epochs to 3. Sampling was not used during inference.

\subsection{Results}
\label{ssec:results}


\input{tables/results_2_chexpert}

We measured performance when trained on MIMIC-CXR, CheXpert Plus, and their combination. Since MAIRA-2 is originally trained on a large mix of private data, we fine-tuned it using Interpret-CXR collection \cite{xu2024overview}. Our results show a consistent improvement in the primary evaluation metric (inverted RadCliQ-v1) across these datasets; see Tab.~\ref{tab:results_mimic} and~\ref{tab:results_chexpert}, with \textbf{*} indicating a statistically significant improvement ($\text{P-value} < 0.05$) under one-sided Wilcoxon signed-rank test. This suggests that leveraging CXR reports with the proposed RDPO approach, as an additional step following pre-training and SFT, enables models to extract additional useful information from texts, leading to better performance.

However, the IU X-Ray dataset presents a distinct challenge. Due to its stylistic differences from MIMIC-CXR and CheXpert, performance there varies significantly. Specifically, when we observe substantial improvements on MIMIC-CXR or CheXpert, the performance on IU X-Ray often declines (Tab.~\ref{tab:results_iu}). This suggests that domain shifts between datasets play a crucial role in model generalization. Incorporating IU X-Ray into training could potentially lead to a more consistent improvement across all three datasets. But the limited size of its training set (7k compared to 200k in the others) poses a major constraint.

\subsection{Radiologist Assessment}
In addition to numerical evaluation on validation subsets, we conducted a small-scale expert evaluation. A board certified radiologist with 15 years of experience reviewed 100 MIMIC-CXR samples processed by the original \textit{stanford-mimic-chexpert} model and our RDPO-aligned model, assessing clinical quality and completeness of reported findings. The RDPO model was preferred in 33 cases, the baseline in 7, with the remaining cases resulting in ties.

\input{tables/results_3_iu}

\subsection{Examples}

In Fig.~\ref{fig:stanford-img} and Fig.~\ref{fig:maira-img} we provide qualitative examples of how RDPO improves clinical accuracy metrics. RDPO writes more details about pathologies present on the study instead of mimiquing standard cliches.

%% file: tables/results_1_mimic.tex
\begin{table}
    \centering
    \caption{RDPO results on the MIMIC-CXR test set.}

    \resizebox{\linewidth}{!}{%
    \begin{tabular}{lccccccc}
        \toprule
        
        Method & Datasets & RDPO & BLEU & Bert-score & Semb-score & Radgraph-comb & RadCliQ-v1 \\

        \midrule

        Stanford & MIMIC-CXR & \xmark & 0.141 & 0.360 & 0.381 & 0.155 & 0.779 \\
        Stanford & MIMIC-CXR & \cmark & 0.140 & 0.360 & 0.409 & 0.159 & \textbf{0.799*} \\
        \midrule
        Stanford & MIMIC$+$CheXpert & \xmark & 0.140 & 0.369 & 0.396 & 0.168 & 0.807 \\
        Stanford & MIMIC$+$CheXpert & \cmark & 0.142 & 0.367 & 0.416 & 0.169 & \textbf{0.820*} \\
        \midrule
        Stanford & CheXpert Plus & \xmark & 0.076 & 0.306 & 0.331 & 0.118 & 0.681 \\
        Stanford & CheXpert Plus & \cmark & 0.074 & 0.306 & 0.336 & 0.121 & \textbf{0.685*} \\
        \midrule
        CSIRO & MIMIC-CXR & \xmark & 0.145 & 0.326 & 0.368 & 0.149 & 0.736 \\
        CSIRO & MIMIC-CXR & \cmark & 0.146 & 0.326 & 0.383 & 0.152 & \textbf{0.746*} \\
        \midrule
        MAIRA-2 & private & \xmark & 0.162 & 0.411 & 0.431 & 0.195 & 0.909 \\
        MAIRA-2 & Interpret-CXR & \cmark & 0.166 & 0.416 & 0.446 & 0.198 & \textbf{0.932*} \\
        
        \bottomrule
         
    \end{tabular}}
    \label{tab:results_mimic}
\end{table}

%% file: tables/results_2_chexpert.tex
\begin{table}
    \centering
    \caption{RDPO results on the CheXpert Plus test set.}

    \resizebox{\linewidth}{!}{%
    \begin{tabular}{lccccccc}
        \toprule
        
        Method & Datasets & RDPO & BLEU & Bert-score & Semb-score & Radgraph-comb & RadCliQ-v1 \\

        \midrule

        Stanford & MIMIC-CXR & \xmark & 0.117 & 0.243 & 0.320 & 0.106 & 0.626 \\
        Stanford & MIMIC-CXR & \cmark & 0.126 & 0.255 & 0.380 & 0.132 & \textbf{0.673*} \\
        \midrule
        Stanford & MIMIC$+$CheXpert & \xmark & 0.127 & 0.286 & 0.417 & 0.179 & 0.746 \\
        Stanford & MIMIC$+$CheXpert & \cmark & 0.147 & 0.295 & 0.464 & 0.190 & \textbf{0.790*} \\
        \midrule
        Stanford & CheXpert Plus & \xmark & 0.122 & 0.279 & 0.385 & 0.160 & 0.711 \\
        Stanford & CheXpert Plus & \cmark & 0.124 & 0.279 & 0.403 & 0.160 & \textbf{0.720} \\
        \midrule
        CSIRO & MIMIC-CXR & \xmark & 0.101 & 0.193 & 0.254 & 0.099 & 0.569 \\
        CSIRO & MIMIC-CXR & \cmark & 0.108 & 0.206 & 0.291 & 0.118 & \textbf{0.598*} \\
        \midrule
        MAIRA-2 & private & \xmark & 0.113 & 0.285 & 0.327 & 0.136 & 0.674 \\
        MAIRA-2 & Interpret-CXR & \cmark & 0.127 & 0.295 & 0.362 & 0.152 & \textbf{0.707*} \\
        
        \bottomrule
         
    \end{tabular}}
    \label{tab:results_chexpert}
\end{table}

%% file: tables/results_3_iu.tex
\begin{table}
    \centering
    \caption{RDPO results on the IU X-Ray test set.}

    \resizebox{\linewidth}{!}{%
    \begin{tabular}{lccccccc}
        \toprule
        
        Method & Datasets & RDPO & BLEU & Bert-score & Semb-score & Radgraph-comb & RadCliQ-v1 \\

        \midrule

        Stanford & MIMIC-CXR & \xmark & 0.208 & 0.433 & 0.592 & 0.173 & \textbf{1.074} \\
        Stanford & MIMIC-CXR & \cmark & 0.212 & 0.433 & 0.578 & 0.181 & 1.070 \\
        \midrule
        Stanford & MIMIC$+$CheXpert & \xmark & 0.207 & 0.466 & 0.587 & 0.212 & \textbf{1.201} \\
        Stanford & MIMIC$+$CheXpert & \cmark & 0.212 & 0.456 & 0.592 & 0.203 & 1.164 \\
        \midrule
        Stanford & CheXpert Plus & \xmark & 0.165 & 0.414 & 0.490 & 0.129 & 0.894 \\
        Stanford & CheXpert Plus & \cmark & 0.174 & 0.419 & 0.509 & 0.138 & \textbf{0.926*} \\
        \midrule
        CSIRO & MIMIC-CXR & \xmark & 0.201 & 0.406 & 0.611 & 0.195 & 1.077 \\
        CSIRO & MIMIC-CXR & \cmark & 0.212 & 0.417 & 0.620 & 0.234 & \textbf{1.171*} \\
        \midrule
        MAIRA-2 & private & \xmark & 0.197 & 0.474 & 0.603 & 0.218 & \textbf{1.255} \\
        MAIRA-2 & Interpret-CXR & \cmark & 0.198 & 0.466 & 0.600 & 0.217 & 1.231 \\
        
        \bottomrule
         
    \end{tabular}}
    \label{tab:results_iu}
\end{table}

%% file: sections/5_conclusion.tex
\begin{figure}
    \centering
    \begin{minipage}[l]{0.3\textwidth}
    \centering
        \includegraphics[width=\linewidth]{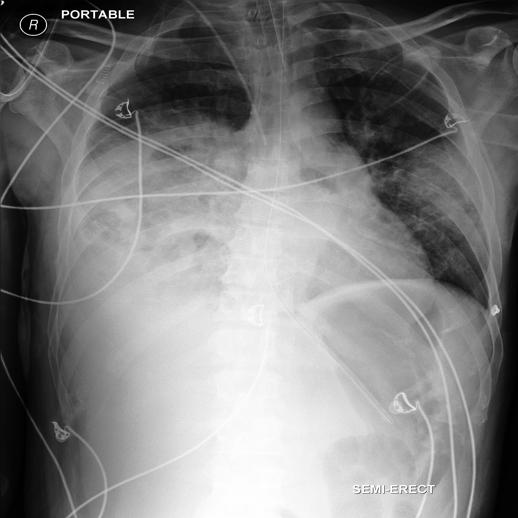}
        \label{fig:sample_figure} 
    \end{minipage}\hfill
    \begin{minipage}[r]{0.65\textwidth}
    \fontsize{8}{10}\selectfont
    \textbf{GT:} As compared to the previous radiograph, the patient has received a new orogastric tube. The course of the tube is unremarkable, the tip of the tube projects over the middle parts of the stomach. There is no evidence of complications. The other monitoring and support devices are in unchanged position. \textit{The massive bilateral lung abnormalities are constant in appearance.} \\
    \textbf{Baseline:} as compared to the previous radiograph, the patient has received a nasogastric tube. the course of the tube is unremarkable, the tip of the tube projects over the middle parts of the stomach. the other monitoring and support devices are constant. \textit{constant appearance of the lung parenchyma} and the cardiac silhouette. \\
    \textbf{RDPO:} endotracheal tube terminates approximately 3.8 cm above the carina. an orogastric tube courses below the diaphragm, tip is not included in this examination. a right internal jugular catheter terminates in the region of the cavoatrial junction. \textit{there is a large right pleural effusion with adjacent atelectasis. there is also a small to moderate right pleural effusion. there is also a small left pleural effusion.} there is no pneumothorax. the heart and mediastinum are within normal limits despite the projection.
    \end{minipage}
    \centering%
    \caption{Test sample from MIMIC-CXR, generated by stanford-mimic-chexpert model. RDPO generates report with more clinical details.}
    \label{fig:stanford-img}
\end{figure}

\begin{figure}
    \centering
    \begin{minipage}[l]{0.3\textwidth}
    \centering
        \includegraphics[width=\linewidth]{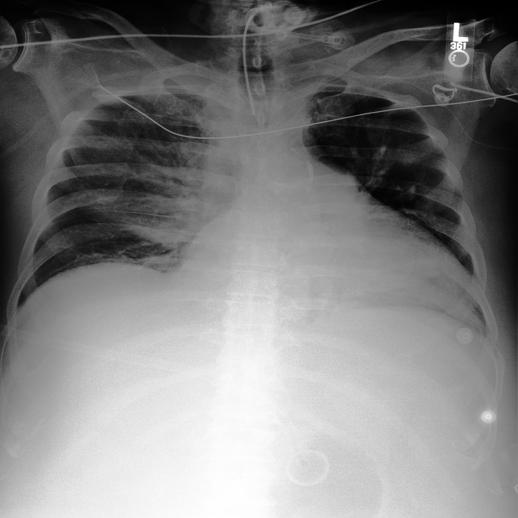}
        \label{fig:sample_figure}
    \end{minipage}\hfill
    \begin{minipage}[r]{0.65\textwidth}
    \fontsize{8}{10}\selectfont
    \textbf{GT:} Single AP portable chest radiograph is obtained.
    Tracheostomy tube is present.
    There is no pneumothorax or pleural effusion.
    \textit{There is a hazy veil-like opacity in the right upper lung zone which may be consolidation, atelectasis or artifact.}
    Heart size appears enlarged; however, this may be technical due to AP view.
    Bony structures are intact. \\
    \textbf{Baseline:} Single frontal view of the chest demonstrates an ET tube extending 3.5 cm above the carina. A left PICC tip is in the mid SVC. Lung volumes are low, accentuating bronchovascular crowding. There is no pneumothorax or large effusion. \textit{There is no confluent consolidation to suggest pneumonia.} There is no pulmonary edema. \\
    \textbf{RDPO:} Single frontal view of the chest demonstrates an ET tube extending 3.7 cm above the carina. An enteric tube extends inferiorly out of view into the stomach. The heart is enlarged, accentuated by low lung volume. Perihilar vascular markings are prominent, consistent with mild edema. \textit{There is retrocardiac opacity, which could represent atelectasis or consolidation.} There is no pneumothorax or large effusion.
    
    \end{minipage}
    \centering
    \caption{Test sample from MIMIC-CXR, generated by MAIRA-2. RDPO generates a more clinically accurate report.}
    \label{fig:maira-img}
\end{figure}

\section{Conclusion}
\label{sec:conclusion}


This work introduces RDPO --- a novel alignment framework for VLMs in the context of RRG problem. In contrast to the existing alignment methods, e.g. DPO, our approach does not rely on additional annotation of human preferences. Instead, RDPO relies just on SFT datasets comprising image-text pairs. In other words, RDPO allows to get more use out of these datasets than a regular SFT. Our experiments demonstrate that applying RDPO to several publicly available SoTA VLMs further increases their RRG quality. We emphasize that these results can be achieved without the cost of additional human annotation.

We believe that our framework can be further improved and scaled over larger data collections, leading to even more fascinating results.  Alternative strategies for sampling a rejected response are also worth studying.